\pdfoutput=1

\documentclass[sigconf,natbib=true]{acmart}

\usepackage{balance}
\usepackage{latexsym}

\usepackage[T1]{fontenc}

\usepackage[utf8]{inputenc}

\usepackage{microtype}

\usepackage{algorithm}
\usepackage{algorithmicx}

\makeatletter
\newcommand\fs@nobottomruled{\def\@fs@cfont{\bfseries}\let\@fs@capt\floatc@ruled
  \def\@fs@pre{\hrule height.8pt depth0pt \kern2pt}%
  \def\@fs@post{}
  \def\@fs@mid{\kern2pt\hrule\kern2pt}%
  \let\@fs@iftopcapt\iftrue}
\makeatother

\floatstyle{nobottomruled}
\restylefloat{algorithm}

\usepackage{listings}
\usepackage{xcolor}

\definecolor{codeblue}{rgb}{0,0.4,0.6}
\lstdefinestyle{mystyle}{
    commentstyle=\color{codeblue},
    basicstyle=\ttfamily\footnotesize,
    breakatwhitespace=false,         
    breaklines=true,                 
    captionpos=b,                    
    keepspaces=true,                 
    numbers=left,                    
    numbersep=5pt,  
    frame=none,
    showspaces=false,                
    showstringspaces=false,
    showtabs=false,                  
    literate={\ \ }{{\ }}1,
    basicstyle=\fontsize{9}{13}\selectfont\ttfamily 
}

\usepackage{adjustbox}
\usepackage{booktabs}
\usepackage{multirow}
\usepackage{multicol}
\usepackage{algpseudocode}
\usepackage{tikz}
\usepackage{pgfplots}
\usepackage{amsmath}
\usepackage{amsfonts}
\usepackage{mathtools}
\usepackage{newfloat}
\usepackage{listings}
\usepackage{enumitem}

\pgfplotsset{compat=1.8}

\floatstyle{ruled}
\newfloat{listing}{tb}{lst}{}
\floatname{listing}{Listing}

\usepackage{xargs}

\title{InceptionXML: A Lightweight Framework with Synchronized Negative Sampling for Short Text Extreme Classification}

\author{Siddhant Kharbanda}
\affiliation{%
  \institution{Aalto University}
  \city{Espoo}
  \country{Finland}}
\affiliation{%
  \institution{Microsoft Corporation}
  \country{India}
}
\email{skharbanda@microsoft.com}

\author{Atmadeep Banerjee}
\affiliation{%
  \institution{Aalto University}
  \city{Espoo}
  \country{Finland}}
\email{atmadeepb@gmail.com}

\author{Devaansh Gupta}
\affiliation{%
  \institution{Aalto University}
  \city{Espoo}
  \country{Finland}
}
\affiliation{%
  \institution{BITS Pilani}
  \city{Pilani}
  \country{India}}
\email{f20190187@pilani.bits-pilani.ac.in}

\author{Akash Palrecha}
\affiliation{%
  \institution{Aalto University}
  \city{Espoo}
  \country{Finland}}
\email{akashpalrecha@gmail.com}

\author{Rohit Babbar}
\affiliation{%
  \institution{Aalto University}
  \city{Espoo}
  \country{Finland}
}
\affiliation{%
  \institution{University of Bath}
  \city{Bath}
  \country{UK}
  }
\email{rohit.babbar@aalto.fi}

\copyrightyear{2023} 
\acmYear{2023} 
\setcopyright{acmlicensed}\acmConference[SIGIR '23]{Proceedings of the 46th International ACM SIGIR Conference on Research and Development in Information Retrieval}{July 23--27, 2023}{Taipei, Taiwan}
\acmBooktitle{Proceedings of the 46th International ACM SIGIR Conference on Research and Development in Information Retrieval (SIGIR '23), July 23--27, 2023, Taipei, Taiwan}
\acmPrice{15.00}
\acmDOI{10.1145/3539618.3591699}
\acmISBN{978-1-4503-9408-6/23/07}

\begin{document}
\begin{abstract}
Automatic annotation of short-text data to a large number of target labels, referred to as Short Text Extreme Classification, has found numerous applications including prediction of related searches and product recommendation. 
In this paper, we propose a convolutional architecture \textsc{InceptionXML} which is light-weight, yet powerful, and robust to the inherent lack of word-order in short-text queries encountered in search and recommendation. 
We demonstrate the efficacy of applying convolutions by recasting the operation along the embedding dimension instead of the word dimension as applied in conventional CNNs for text classification.
Towards scaling our model to datasets with millions of labels, we also propose \textsc{SyncXML} pipeline which improves upon the shortcomings of the recently proposed dynamic hard-negative mining technique for label shortlisting by synchronizing the label-shortlister and extreme classifier.
\textsc{SyncXML} not only reduces the inference time to half but is also an order of magnitude smaller than state-of-the-art \textsc{Astec} in terms of model size. 
Through a comprehensive empirical comparison, we show that not only can \textsc{InceptionXML} outperform existing approaches on benchmark datasets but also the transformer baselines requiring only 2\% FLOPs. The code for \textsc{InceptionXML} is available at \url{https://github.com/xmc-aalto/inceptionxml}.
\end{abstract}



\maketitle
\section{Introduction}
\begin{figure}[th]
\includegraphics[width=\columnwidth]{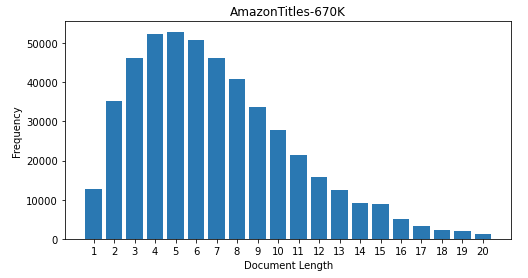}
\label{fig:wiki325}
\caption{Input sequence length plotted against frequency for AmazonTitles-670K dataset. For this dataset, 70\% of training instances have $\leq$ 8 words, and 30\% have $\leq$ 4 words.}
\label{fig:length}
\end{figure}

Extreme Multi-label Classification (\textbf{XML}) involves classifying instances into a set of most relevant labels from an extremely large (on the order of millions) set of all possible labels \cite{Agrawal13}. 
By mapping search queries and items (resp.) to labels, it has been shown in recent works that ranking and recommendation (resp.) tasks, such as -- prediction of related searches on search engines \cite{Jain19}, suggestion of query phrases corresponding to short textual description of products on e-stores \cite{Chang20}, product-to-product recommendation \cite{Dahiya21b, chang2021extreme}, query-to-bid matching \cite{Dahiya21} and, dynamic search ads \cite{Mittal21b} can be reformulated as \textit{XML problems with short-text inputs}. 
Short-text inputs encountered in search and recommendation consist of input queries with 3-8 tokens on average. These observations are in line with the statistics of public datasets used in XML\footnote{http://manikvarma.org/downloads/XC/XMLRepository.html} (Ref: \autoref{fig:length} \& \autoref{tbl:datasets}). 
Unlike the standard text classification problem involving long documents, short text input data leads to additional challenges. \\ 

\begin{figure}[htp]
    \centering
    \includegraphics[width=1.05\columnwidth]{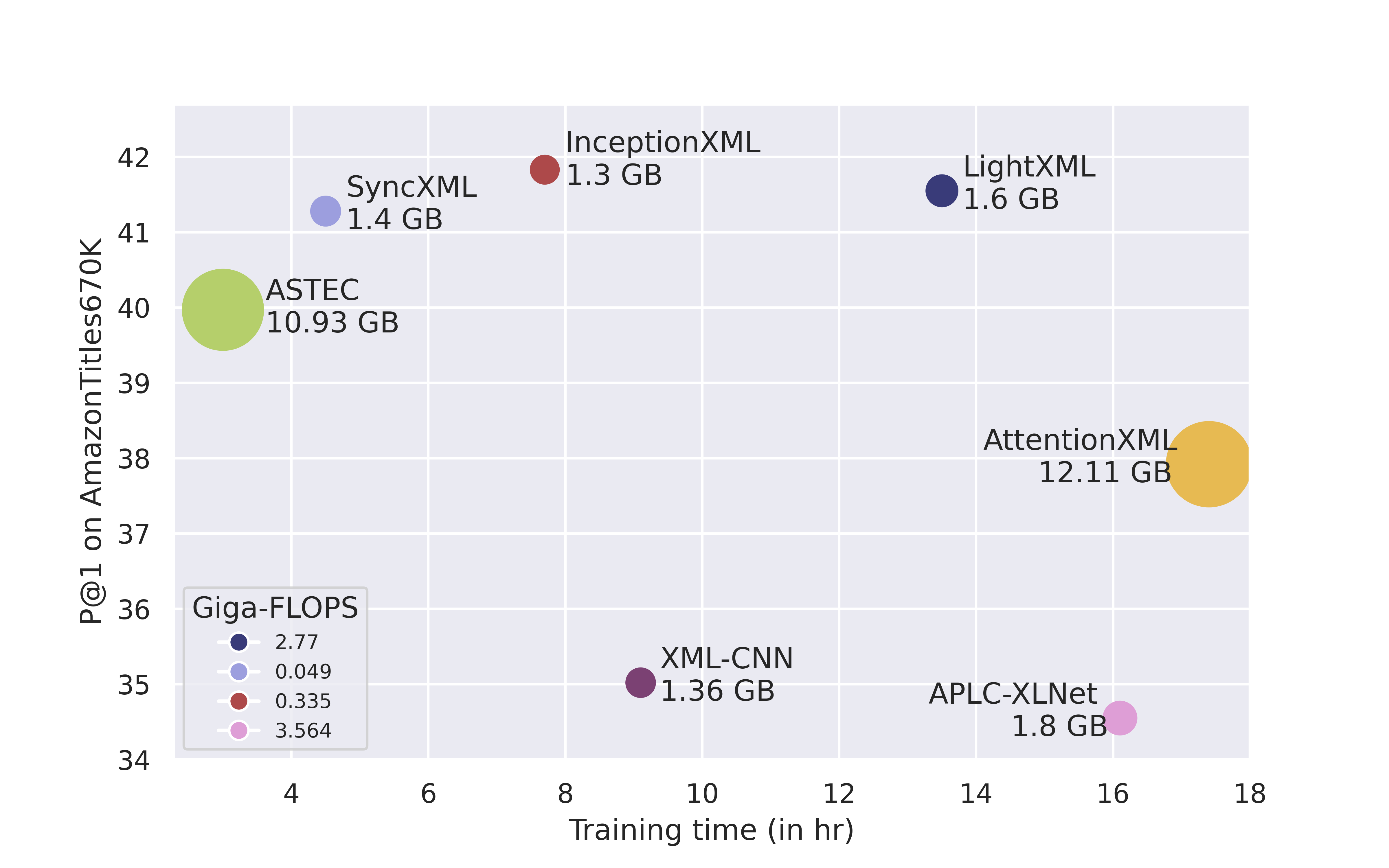}
    \caption{\textsc{SyncXML} \textit{(Ours)} hits the sweet spot in terms of scores on the P@1 metric, training time and model size.}
    \label{fig:efficient}
\end{figure}

\noindent \textbf{Challenges in Short-Text XML} Firstly, the short-text input queries are sparse and contain very few words. 
Secondly, these are plagued with noise and non-standard phrases which do not always observe the syntax of a written language. 
For instance, queries ``\textit{best wireless headphones 2022}" and ``\textit{2022 best headphones wireless}" should invoke similar search results on an e-commerce website \cite{tayal2020regularized, wang2016understanding}.
Thirdly, a large fraction of classes are tail labels, which are paired with a handful of positive samples \cite{Jain16, qaraei2021convex}. 
Taken together, the above characteristics, pose a challenge in learning rich feature representations for the task at hand.
In this paper, to mitigate the aforementioned problems we develop \textbf{InceptionXML} - a novel CNN-based encoder which derives its motivation from two key observations. \\

\noindent \textbf{Motivation 1: Need of lightweight architectures in Short-Text XML:}
While large pre-trained language models are the default choice for most down-stream language tasks, in this paper, we argue that (i) using such computationally intensive architectures for modeling short-text queries is rather overcompensating, especially for the XML task at hand.
Further, (ii) the real-world use cases of short-text extreme classification require very fast inference times. The deployment of large pre-trained language models as XML encoders \cite{Jiang2021, Ye20, Chang20} adds heavily to the existing compute costs in XML tasks leading to slow training and inference times (\autoref{tbl:computation_stats}). 
Finally, (iii) extremely large number of possible labels leads to memory bottlenecks in XML tasks \cite{Chang20}. As a result, these transformer-based methods become unscalable to labels in the order of millions (\autoref{tbl:results}) while staying within reasonable hardware constraints. \\

\noindent \textbf{Motivation 2: Drawbacks of conventional CNNs in short-text classification:} 
Traditionally, in the usage of CNNs over words in text classification, the intent is to capture the occurrences of $n$-grams for representation learning \cite{Kim14, Liu17, conv_knrm}. While \cite{Kim14, Liu17} simply leverage convolutional filters with filter sizes equal to the embedding dimension and window sizes spanning multiple n-gram lengths, \cite{conv_knrm} goes a step further and soft-matches the obtained n-gram features between documents, thus promoting retrieval on the basis of multiple concepts in the document. 
We argue that this formulation is unsuitable for short-text classification problems as (i) the implicit but incorrect assumption of proper \textit{word-ordering} in short-text queries \cite{wang2016understanding} for efficient n-gram capture, and (ii) as explained next, the much \textit{smaller sequence length} that restricts the effectiveness of convolution in CNNs over the inputs. 
In the datasets derived from Wikipedia titles, $98\%$ documents have 8 or less words, while $82\%$ have $4$ words or less (\autoref{tbl:datasets}). 
Moreover, $70\%$ of the instances in AmazonTitles-670K consist of $8$ words or less (\autoref{fig:length}).
This makes the convolutional filters spanning over 4-8 words in \cite{Kim14, Liu17, Wang2017CombiningKW} behave analogously to a weak fully connected layer with very few hidden units, and hence leading to sub-optimal feature maps for representation learning.
\\

\noindent \textbf{Contributions: } These insights motivate us to develop a lightweight architecture and investigate the suitability of \textit{convolving over the embedding dimensions} of the input in contrast to the words for short-text queries. 
We thus (i) develop \textsc{InceptionXML}, a lightweight CNN-based encoder, which goes against the traditional paradigm \cite{Kim14,Liu17,conv_knrm} of convolving over the words dimension in favor of the embedding dimension, (ii) propose an embedding-enhancement module for learning a word-order agnostic representation of the input, making our approach more robust to lack of structure in short-text queries.
Towards scaling the encoder to labels in the order of millions, we (iii) highlight and mitigate the drawbacks of dynamic hard-negative mining by creating a fast and computationally inexpensive model training pipeline \textsc{SyncXML} 
which synchronizes the label-shortlisting and extreme classification tasks.
Finally, we (iv) do a comprehensive comparison of our proposed InceptionXML encoder with pre-trained transformer baselines, showing transformers are perhaps an overkill for the short-text XML task.\\

\noindent \textbf{Highlights:} We (i) further the state-of-the-art on 23 out of 24 metrics across 4 popular benchmark datasets (ii) reduce the inference time to half of the previous fastest state-of-the-art, and (iii) require only $2\%$ FLOPS as compared to pre-trained transformer based approaches. 
\noindent A pictorial Vis-à-vis comparison of InceptionXML against state-of-the-art approaches 
on metrics like accuracy, model size, training/inference times 
is shown in \autoref{fig:efficient}.

\section{Related Work}

\noindent \textbf{Extreme Classification:} The majority of initial works focused on designing one-vs-rest \cite{Babbar17,Babbar19, schultheis2022speeding}, tree-based \cite{Prabhu18b, chalkidis2019large, Khandagale19} or label embedding based \cite{Bhatia15} classifiers with fixed features in the form of bag-of-words representation.  
With advances in deep learning, jointly learning label and input text spaces has also been developed \citep{tang2015pte, wang2018joint}.
For XML tasks, recent techniques based on attention mechanism \cite{you18} and pre-trained transformer models \cite{Chang20, Ye20, Jiang2021, yu2020pecos} have shown great promise.  
For text classification, while \cite{Wang2017CombiningKW} extended \cite{Kim14} for short input sequences, \cite{Liu17} built upon it for XML tasks. \\

\noindent \textbf{Short-text Extreme Classification:} In XML tasks where the inputs are short text queries, there has been a slew of recent works.
Based on the availability of label meta-data, these works can be divided into two categories: (i) ones which make no assumptions regarding label text, i.e., labels are numeric identifiers, such as \textsc{Astec} \cite{Dahiya21} and (ii) others which assume that the labels are endowed with clean label text which include \textsc{Decaf} \cite{Mittal21}, \textsc{GalaXC} \cite{Saini21}, \textsc{Eclare} \cite{Mittal21b}, and \textsc{SiameseXML} \cite{Dahiya21b}.
Even though the additional label meta-data is useful, it is usually only known for only a small subset of all labels. 
For instance, on AmazonTitles-3M dataset, label text is available only for 1.3 million labels.
Further, the former problem setup, which is the focus of this work, makes no assumption about label-text, and hence is harder, more general and widely applicable.\\

\noindent \textbf{Dense Retrieval (DR) and XML:} 
Most recent DR approaches aimed at solving the question-answering task \cite{rocketqav1, ernie_search, colbert, ance, zhang2021adversarial} leverage two-tower models that embed short-text queries/questions and long-text documents containing answers into a common embedding space using pre-trained transformer models. These encoders are trained to solve a retrieval and re-ranking task. Due to the aforementioned computational constraints in XML, both problems principally end up sharing the retrieval task. 
However, XML methods make use of a ``label shortlister'' or a ``meta-classifier'' that helps retrieve the K most probable labels that are used to train the extreme classifier, which replaces traditional neural re-ranking task in dense retrieval. 
Further, while DR methods show promise, they are not adaptable to the paper's set up, due to the absence of label-texts. \\ 

\noindent \textbf{Negative Sampling:} Given the scale of the XML problem, negative label sampling becomes an imperative task. A popular approach adapted in DR \cite{dpr, ance, realm} and XML works alike is to make use of an ANNS graph. In short-text XML, works following the DeepXML pipeline \cite{Dahiya21b, Mittal21, Mittal21b, Saini21} shortlist negatives by first training a frugal encoder on label-clusters, and then use an ANNS graph \cite{Malkov2020EfficientAR} to create a static shortlist of hard-negatives.
Meanwhile, other XML works like \cite{Jiang2021, cascadeXML} propose an alternate \textit{dynamic} negative mining solution which enables end-to-end learning of the label-shortlister and extreme classifiers, and leverages the model's current weights to determine the hard negatives for extreme classifier.
\begin{figure*}[htp]
    \centering
    \includegraphics[width=0.93\textwidth]{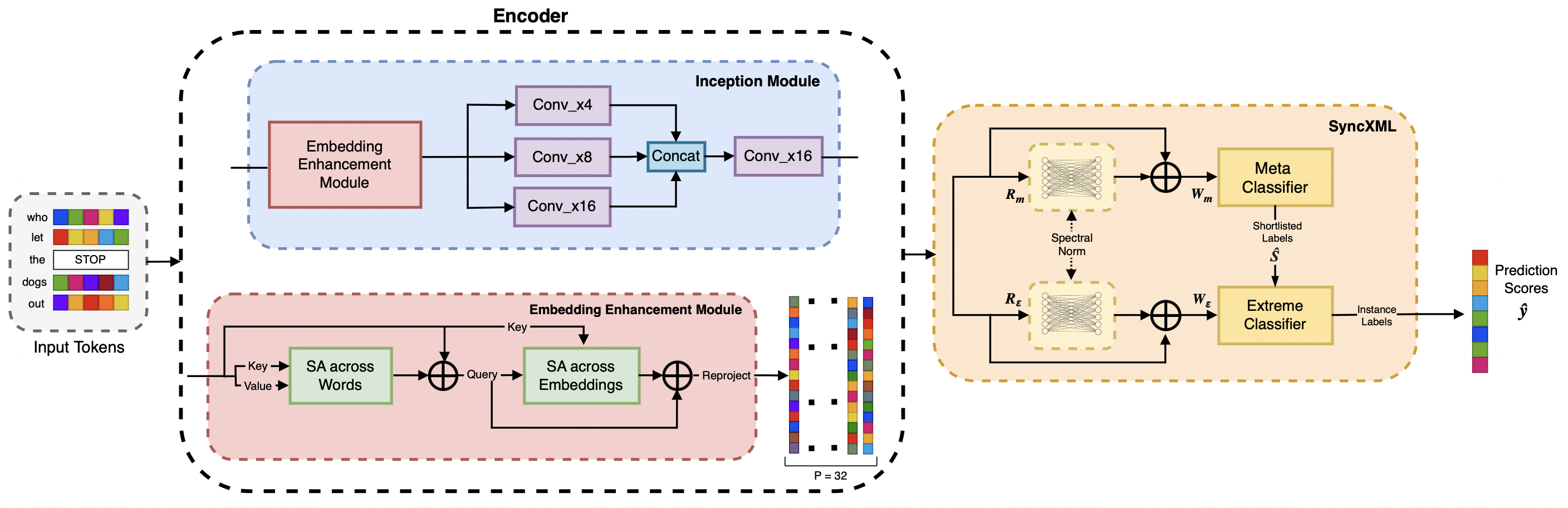}
    \caption{The plug-n-play \textsc{SyncXML} training pipeline demonstrated with InceptionXML as an encoder.}
    \label{fig:model}
\end{figure*}

\section{Proposed Encoder - \textsc{I\lowercase{nception}XML}}

\noindent \textbf{Problem Setup:} Given a training set $\{x_i, y_i\}_{i=1}^N$, $x_i$ represents an input short-text query, and the corresponding label set is represented by $y_i \in \{0, 1\}^L$ where $L$ denotes the total number of labels. 
It may be noted that even though $L\sim 10^6$, an instance is only annotated with a few positive labels (Table: \ref{tbl:datasets}).
The goal is to learn a classifier which, for a novel test instance $x'$, predicts the top-k labels towards better precision@k and propensity-scored precision@k \cite{Bhatia16} metrics. Towards this goal, the main body of our encoder consists of three modules that are applied sequentially on the word embeddings (Fig. \ref{fig:model}). These are (i) an embedding enhancement module, (ii) embedding convolution layers and (iii) an extreme linear classifier. 
\subsection{Embedding Convolutions}
Models generating word embeddings \cite{Pennington2014GloVeGV, Mikolov13, joulin2016bag} have been successful in capturing fine-grained semantic and syntactic regularities, especially in the analogy reason task, using vector arithmetic \cite{Pennington14}. Word embeddings capturing local or global context vectors have shown to place related words near each other in the embedding space with certain directions in space corresponding to semantic distinctions, thus producing dimensions of meaning \cite{Pennington14}. Not limited to this, further grammatical information such as number and tense are also represented via directions in space \cite{NEURIPS2019_159c1ffe}.

In such a paradigm where individual embedding dimensions capture a fine-grained semantic, syntactic regularity or tense, we hypothesize that embedding convolutions tend to capture ``coupled semantics'' across different dimensions by convolving over these word embeddings in a stacked setting. Further, embedding convolutions seek to find similar patterns across these coupled semantics by processing a limited subset of dimensions at a time.  As compared to traditional convolutional operation, embedding convolutions create significantly larger and enriched activation maps for the same inputs, while requiring substantially lesser parameters by using smaller filters of size  $\mathbb{R}^{S \times 16}$, where $S$ is the maximum sequence length of the input.
We show empirically, that this modified approach works well for both \textit{short} as well as \textit{medium} queries of up to $32$ words, significantly outperforming conventional CNN-based approaches \cite{Liu17, Kim14} for short-text XML task. 

As some readers might rightfully argue, pre-trained word embeddings are typically not trained with any incentive for localizing complementary semantic information in the adjacent embedding dimension. 
More specifically, the coupled semantics that our model tries to extract may initially occur across spatial distances that are too large for the convolution kernel to detect.
To this end, we process the stacked word embeddings with attention based embedding enhancement module before applying embedding convolutions. This lets information flow across every pair of semantics irrespective of the spatial distance between them.

\subsection{Embedding Enhancement Module}
As mentioned in section 1, the sequence of words in the short-text extreme classification setting do not necessarily observe the syntax of written language, and the presence of a word is more important than its position \cite{wang2016understanding}.
Therefore, creating a model agnostic to the input word order is desirable.
There exist pooling techniques to make the input word order agnostic, for example, \cite{Dahiya21} use a tf-idf weighted sum of word embeddings as an input to \textsc{Astec}.
However, pooling, especially max pooling, often results in loss of information.
In a paradigm, where the input is a sequence of 300-dimensional embeddings of 3-5 words, the information is already very scarce and needs alternate methods to be made word order agnostic. 
To alleviate this, we propose an embedding enhancement module. 

Specifically, the module consists of two orthogonal attention layers applied sequentially on the word and the embedding dimensions followed by a projection layer, effectively encoding global information both, on a word-level and on a semantic-level (Figure \ref{fig:SA-ablation}).
The sequential attention formulation in our embedding enhancement module is given by: 
\begin{gather}
    x_{sa} = \textrm{Attn}(q = E(x),\; k = E(x),\; v = E(x)) \label{eqn:saw} \nonumber \\
    x_{sa} = \textrm{Attn}(q = x_{sa}^T,\; k = E(x)^T,\; v = E(x)^T)^T \nonumber
\end{gather}
where $E(x)$ denotes the stacked word embeddings for a sample text input $x$ such that $E(x) \in \mathbb{R}^{S \times d}$.
Finally, each dimension of the intermediate embeddings $x_{sa}$ is then projected to a $p$-dimensional space where $p = 32$ to obtain the final enhanced embeddings 
$x_{enh} \in\ \mathbb{R}^{p \times d}$.
The information flow across the embeddings in this module followed by per-dimension projection makes $x_{enh}$ independent of the word order in short-text queries and makes our model more robust to their lack of structure.

\subsection{Embedding Convolution Layers}
Motivated by the popular InceptionNet architecture \cite{InceptionNet}, we employ three parallel branches of 1D convolution layers $V_i,\, i \in [1,2,3]$ with filter sizes of $w_i$ where $w_i \in [4, 8, 16]$ each with a stride of $4$ along the embedding dimension and $p$ output channels. Let $h_{w_i}$ be the result of applying $V_i$ over $x_{enh}$. We concatenate all resultant $h_{w_i}$ row-wise before passing them to the next layer.
\begin{eqnarray}\label{eq:conv}
     h_{w_i} & =  & V_i * x_{enh} \nonumber \\
     h_f &  =  & V_f * [h_{w_1}, h_{w_2}, h_{w_3}] 
\end{eqnarray}
\noindent A final embedding convolutional layer $V_f$ with kernel size of 16 and stride 4 is applied on the concatenated feature map, which is further flattened to form the final feature representation $h_f$. 
This formulation allows $V_f$ to have an effective receptive field spanning $1/4^{th}$ of the enhanced embeddings, further obviating the locality constraints of CNNs as highlighted in section 3.

\subsection{Extreme Linear Classifier}

The first layer $R$ transforms the feature map from the encoder with a skip-connection while keeping the dimensions same. The next linear layer $W$ has one-vs-all classifiers for each label in the dataset which projects the features to the label space.
\begin{equation}
    \hat{y} = \sigma(W \cdot (relu(R \cdot h_f) + h_f)) \nonumber
\end{equation}

\noindent The model is trained end-to-end using binary cross entropy loss.  
\begin{equation} 
    \textrm{BCE}(y, \hat{y}) = -\sum_{j \in L} (1 - y_j)\log(1 - \hat{y}_j) + y_j\log(\hat{y}_j) \nonumber
\end{equation}

\section{S{\small ync}XML Training Pipeline}
\textsc{InceptionXML} described previously scales to datasets with hundreds of thousands of labels. However, scaling up to millions of labels in its existing form is difficult as the loss computation in equation above involves calculation of loss over all $L$ labels, a very large majority of which are negative labels for a given instance. 
Even with sufficient hardware resources, scaling up over the entire label space requires exceedingly longer training times \cite{Chang20}. 
We thus propose the \textsc{SyncXML} training pipeline, which improves existing hard-negative mining to enable scaling to output spaces in the order of millions along with an updated training schedule. Not only does the pipeline scale our encoder, but also significantly reduces the training time and computational cost (\autoref{tbl:model_stats}). \\

\noindent \textbf{Hard Negative-Mining of Labels:}
While XML methods have made efficient use of ANNS for hard-negative label mining under \textit{fixed representation} of data points \cite{Jain19, Dahiya21}, only limited works \cite{Jiang2021, cascadeXML} explore \textit{dynamic} hard negative-mining which provide hard negatives as per the model's current weights.
Similarly, works in dense retrieval \cite{dpr, ance, realm} propose to learn retrievers using question-document matching algorithms by jointly learning their embeddings and then leverage an ANNS \cite{faiss} to create a hard-negatives shortlist. However, due to the static nature of the ANNS graph, the model ends up using stale embeddings to create this shortlist. To mitigate this, \cite{realm, ance} propose an asynchronous ANNS refresh strategy, which adds significantly to the computational cost. 
Following the approach popularized by recent methods making use of dynamic hard-negative mining, our model makes predictions in two synchronous stages: (i) shortlisting top $K$ label-clusters or ``meta-labels" using a meta-classifier, and (ii) employing a computationally feasible number of one-vs-all classifiers corresponding to the labels included in the shortlisted clusters to get the final predicted labels and perform backpropagation.\\

\noindent \textbf{Label Clustering:} To perform label clustering, we construct Hierarchical Label Tree (HLT) using the labels' Positive Instance Feature Aggregation (PIFA) representation over sparse BOW features of their training samples \cite{Chang20, Dahiya21}. 
Specifically, we use balanced 2-means clustering to recursively partition the label set until we have a mapping $C$ from $L$ labels to $L'$ label clusters where $L' \ll L$ (Table:\ref{tbl:datasets}). 

\begin{figure}[htp]
\centering
\scalebox{0.8}{
\begin{tikzpicture}
        \begin{axis}[
        xlabel={Epochs},
        ylabel={Training P@1},
        grid=major, no markers, thick, 
        legend cell align={left}, legend style={font=\footnotesize,at={(0.99,0.01)},anchor=south east},
        xmin=-1,xmax=48,ymin=-1,ymax=100
        ]
            \addplot+[color=blue,dashed] table {stats-1.txt};
            \addplot+[color=red,dashed] table {stats-2.txt};
            \addplot+[color=blue] table {stats-3.txt};
            \addplot+[color=red] table {stats-4.txt};
            \legend{LightXML extreme, LightXML meta, SyncXML extreme, SyncXML meta}
        \end{axis}
    \end{tikzpicture}
}
\vspace{-2mm}
\caption{Progress of training (Precision@1) for the extreme and meta-classifier of \textsc{LightXML} and \textsc{SyncXML} training pipelines on AmazonTitles-670K}
\label{fig:GroupTrain}
\end{figure}

\noindent \textbf{Drawbacks of \textsc{LightXML} framework:} 
From \autoref{fig:GroupTrain} we observe that the performance of our encoder is bottlenecked by a poorly performing meta-classifier. While we witness a smooth increment in the training P@1 values for the extreme classifier (dashed blue), the meta-classifier is unable to catch-up (dashed red).
This indicates that the meta-classifier used in dynamic hard-negative mining is not aligned well enough for the encoder to learn suitable representations to provide the extreme classifier with suitable hard-negatives. 
Our observations also indicate the fact that the extreme task is easier to learn given a label shortlist, and the model tends to learn representations that benefit the extreme task at the expense of the meta-task. 
\textsc{SyncXML} mitigates these challenges in two ways:
(i) we propose architectural modifications to enable the encoder to learn representations mutually benefitting both tasks, and 
(ii) adjust the training loop to warm start the meta-classifier training while keeping the encoder in sync with the extreme task.
\subsection{Synchronized Architecture}
To synchronize the training of extreme and meta-classifier tasks, we give them similar structures by adding a linear layer $W_m$ with a residual connection $R_m$ before the meta-classifier. Using the intermediate representation $h_f$ from equation (\ref{eq:conv}), this gives :
\begin{equation}
    \hat{y}_m = \sigma(W_m \cdot (relu(R_m \cdot h_f) + h_f))
    \nonumber
\end{equation}

\noindent We create a shortlist $\hat{\mathcal{S}}$ of all the labels in the top $K$ label clusters as predicted by the meta-classifier using a label cluster to label mapping $C^{-1}$. 
Via the linear mapping $W_e$, extreme classifier then predicts the probability of the query belonging to only these shortlisted labels, instead of all $L$ labels.
\begin{align}
    \hat{\mathcal{S}} &= C^{-1}(\ top_K(\hat{y}_m, k)\ ) \nonumber \\
    g_e &= relu(R_{e} \cdot h_f) + h_f \nonumber\\
    \hat{y}_{e,l} &= \sigma(W_{e, l} \cdot g_e), \  \forall l \in \hat{\mathcal{S}} \nonumber
\end{align}
Naturally, architectural similarity alone is not enough to ensure synchronized representation learning. To help the encoder learn representations benefitting both tasks, we further sync them by (i) increasing the ``extremeness'' of the meta-task by enlarging the fan out of label clusters, and (ii) preventing the drift of encoder representations in both classifiers by applying spectral norm \cite{Dahiya21} on the penultimate linear layers of both heads initialized with the same weights.
Not only does this heavily improve (Table: \ref{tbl:ablation}) upon the original implementation of dynamic negative-hard mining as proposed in \cite{Jiang2021}, but also inherently combines the task of all the four stages of the DeepXML pipeline \cite{Dahiya21} into an end-to-end training pipeline.
\begin{table*}[!t]
\centering
\small
\begin{adjustbox}{width=0.9\textwidth,center}
\scalebox{0.95}{
\begin{tabular}{c|c|c|c|c|c|c|c|c}
\toprule
\textbf{Datasets} & \textbf{\# Features} & \textbf{\# Labels} & \textbf{\# Training} & \textbf{\# Test} & \textbf{APpL} & \textbf{ALpP} & \textbf{\#W$\leq4$} & \textbf{\#W$\leq8$}\\
\midrule
\textbf{WikiSeeAlsoTitles-350K} & 91,414  & 352,072 & 629,418   & 162,491 & 5.24  & 2.33  & 82\% & 98\%\\
\textbf{WikiTitles-500K} & 185,479 & 501,070 & 1,699,722 & 722,678 & 23.62 & 4.89  & 83\% & 98\% \\
\textbf{AmazonTitles-670K} & 66,666  & 670,091 & 485,176   & 150,875 & 5.11  & 5.39 & 40\% & 70\% \\
\textbf{AmazonTitles-3M} & 165,431 & 2,812,281 & 1,712,536 & 739,665 & 31.55 & 36.18 & 15\% & 52\% \\
\midrule
\end{tabular}
}
\end{adjustbox}
\caption{Dataset Statistics. APpL denotes the average data points per label, ALpP the average number of labels per point. \#W is the number of words in the training samples.
}\label{tbl:datasets}
\end{table*}

\begin{algorithm}
\caption{Training Schedule for \textsc{SyncXML}}\label{alg:cap}
\end{algorithm}
\begin{lstlisting}[language=Python]
# x: input token ids, y: true labels
# E: Embeddings Weight Matrix
for epoch in (1, epochs):
    for x, y in data:
        z = E(x)
        h = encoder(z)
        y_meta = meta_classifier(h)
        y_cluster = label_to_cluster(y)
        meta_loss = bce(y_meta, y_cluster)
        
        # shortlisting top K clusters
        top_k = get_top_K_clusters(y_meta, k)
        candidates = cluster_to_label(top_k)
        # add missing positive labels
        candidates = add_missing(candidates,y) 
        
        # detached training
        if epoch <= epochs/4:
            h = h.detach()
        y_ext = ext_classifier(h, candidates)
        ext_loss = bce(y_ext, y, candidates)
        loss = meta_loss + ext_loss
        loss.backward()
        
        # gradient descent
        update(E, encoder, meta_classifier, ext_classifier)
\end{lstlisting}
\par\noindent\rule{\linewidth}{0.4pt}
\subsection{Detached Training Schedule} 
To warm start the meta-classifier training, we detach i.e. stop the flow of gradients from the extreme classifier to the encoder (Algorithm \ref{alg:cap}), for the initial 25\% of the training. This results in shortlisting of suitable hard-negative labels for the extreme classifier to learn during training time and ensures higher recall during inference time (Table: \ref{tbl:ablation}).
Detaching, instead of simply removing the extreme classifier from training, enables the classifier to continuously adapt to the changing encoder representations without allowing it to affect the training of the meta-classifier.
The spectral norm applied to the weights of the penultimate layers in both the heads ensures that the encoder representations learnt for the meta-task remains relevant for the extreme task when its gradients are re-attached.

\noindent \textbf{Loss:} The total loss is obtained as $\mathcal{L} = \mathcal{L}_{meta} + \mathcal{L}_{ext}$, where
\begin{eqnarray*} 
    \mathcal{L}_{meta} &=& \textrm{BCE}(y_m\ ,\ \hat{y}_m),  \nonumber \\ 
    \mathcal{L}_{ext} &=& \textrm{BCE}(y_{e, l}\ ,\ \hat{y}_{e,l})\ \ \forall l \in \hat{\mathcal{S}} \nonumber 
\end{eqnarray*}

\section{Experiments}

\begin{table*}[!htb]
\begin{adjustbox}{width=0.93\textwidth,center}
\begin{tabular}{c|ccc|ccc|ccc|ccc}
\toprule
\textbf{Method} & \textbf{P@1} & \textbf{P@3} & \textbf{P@5} & \textbf{PSP@1} & \textbf{PSP@3} & \textbf{PSP@5} & \textbf{P@1} & \textbf{P@3} & \textbf{P@5} & \textbf{PSP@1} & \textbf{PSP@3} & \textbf{PSP@5}\\

\specialrule{0.70pt}{0.4ex}{0.65ex}
& \multicolumn{6}{c|}{\textbf{AmazonTitles-670K}} & \multicolumn{6}{c}{\textbf{WikiSeeAlsoTitles-350K}}\\
\midrule
\textsc{SyncXML*} & \underline{42.17} & \underline{38.01} & \underline{34.86} & 27.92 & \underline{31.07} & \underline{33.72} & \underline{21.37} & \underline{15.00} & \underline{11.84} & \underline{10.64} & \underline{12.79} & \underline{14.69}\\
\textsc{InceptionXML*} & \textbf{42.45} & \textbf{38.04} & 
\textbf{34.68} & \textbf{28.70} & \textbf{31.48} & \textbf{33.83} & \textbf{21.87} & \textbf{15.48} & \textbf{12.20} & \textbf{11.13} & \textbf{13.31} & \textbf{15.20}\\
\textsc{Astec*} & 40.63 & 36.22 & 33.00 & \underline{28.07} & 30.17 & 32.07 & 20.61 & 14.58 & 11.49 & 9.91 & 12.16 & 14.04 \\
\textsc{Parabel*} & 38.00 & 33.54 & 30.10 & 23.10 & 25.57 & 27.61 & 17.24 & 11.61 & 8.92 & 7.56 & 8.83  & 9.96\\
\textsc{Bonsai*}  & 38.46 & 33.91 & 30.53 & 23.62 & 26.19 & 28.41 & 17.95 & 12.27 & 9.56 & 8.16 & 9.68  & 11.07\\
\textsc{InceptionXML} & 41.78 & 37.47 & 34.15 & 28.17 & 30.96 & 33.31 & 21.54 & 15.19 & 11.97 & 10.93 & 13.05 & 14.92\\
\textsc{Astec}  & 39.97 & 35.73 & 32.59 & 27.59 & \underline{29.79} & 31.71 & 20.42 & 14.44 & 11.39 & 9.83 & 12.05 & 13.94\\
\textsc{LightXML} &  41.57 &  37.19 & 33.90 &  25.23 &  28.79 &  31.92 &  21.25 &  14.36 &  11.11 & 9.60 &  11.48 & 13.05 \\
\textsc{APLC-XLNet} & 34.87 & 30.55 & 27.28 & 20.15 & 21.94 & 23.45 & 20.42 & 14.22 & 11.17 & 7.44 & 9.75 & 11.61\\
\textsc{AttentionXML}  & 37.92 & 33.73 & 30.57 & 24.24 & 26.43 & 28.39 & 15.86 & 10.43 & 8.01 & 6.39 & 7.20  & 8.15\\
\textsc{XML-CNN} & 35.02 & 31.37 & 28.45 & 21.99 & 24.93 & 26.84 & 17.75 & 12.34 & 9.73  & 8.24 & 9.72  & 11.15\\  
\textsc{DiSMEC}  & 38.12 & 34.03 & 31.15 & 22.26 & 25.45 & 28.67 & 16.61 & 11.57 & 9.14 & 7.48 & 9.19  & 10.74\\  

\specialrule{0.70pt}{0.4ex}{0.65ex}
& \multicolumn{6}{c|}{\textbf{AmazonTitles-3M}} & \multicolumn{6}{c}{\textbf{WikiTitles-500K}} \\
\midrule 
\textsc{SyncXML*} & \underline{46.95} & \textbf{45.28} & \textbf{43.45} & \textbf{16.02} & \textbf{18.94} & \textbf{21.03} & 46.64 & \underline{26.74} & \underline{19.06} & \underline{19.98} & \underline{20.40} & \underline{20.63} \\
\textsc{InceptionXML*} & - & - & - & - & - & - & \textbf{48.35} & \textbf{27.63} & \textbf{19.74} & \textbf{20.86} & \textbf{21.02} & \textbf{21.23} \\
\textsc{Astec*} & \textbf{47.64} & \underline{44.66} & \underline{42.36} & \underline{15.88} & \underline{18.59} & \underline{20.60} & 46.60 & 26.03 & 18.50 & 18.89 & 18.90 & 19.30 \\
\textsc{Parabel*} & 46.42 & 43.81 & 41.71 & 12.94 & 15.58 & 17.55 & 42.50 & 23.04 & 16.21 & 16.55 & 16.12 & 16.16\\
\textsc{Bonsai*}  & 46.89 & 44.38 & 42.30 & 13.78 & 16.66 & 18.75 & 42.60 & 23.08 & 16.25 & 17.38 & 16.85 & 16.90\\
\textsc{InceptionXML} & - & - & - & - & - & - & 47.62 & 27.23 & 19.44 & 20.75 & 20.87 & 21.06 \\
\textsc{Astec} & 46.68 & 43.21 & 41.33 & 14.92 & 17.84 & 19.52 & 46.01 & 25.62 & 18.18 & 18.62 & 18.59 & 18.95\\
\textsc{LightXML} & - & - & - & - & - & - & \underline{47.17} & 25.85 & 18.14 & 17.64 & 17.54 & 17.50 \\
\textsc{APLC-XLNet} & - & - & - & - & - & - & 43.56 & 23.01 & 16.58 & 14.73 & 13.19 & 13.47\\
\textsc{XML-CNN} & - & - & - & - & - & - & 43.45 & 23.24 & 16.53 & 15.64 & 14.74 & 14.98\\ 
\textsc{AttentionXML} & 46.00 & 42.81 & 40.59 & 12.81 & 15.03 & 16.71 & 42.89 & 22.71 & 15.89 & 15.12 & 14.32 & 14.22\\
\textsc{DiSMEC}  & 41.13 & 38.89 & 37.07 & 11.98 & 14.55 & 16.42 & 39.89 & 21.23 & 14.96 & 15.89 & 15.15 & 15.43\\
 \bottomrule
\end{tabular}
\end{adjustbox}
\caption{Comparison of InceptionXML to state-of-the-art algorithms on benchmark datasets. The best-performing approach is in \textbf{bold} and the second best is \underline{underlined}. (*) represents results of an ensemble of 3 models.
The algorithms omitted in AmazonTitles-3M do not scale for this dataset on 1 Nvidia V100 GPU.
}
\label{tbl:results}

\end{table*}
\noindent \textbf{Implementation Details:} For tokenization, we use a word-piece tokenizer which is learnt on the training data. 
For the tokens present in our vocab and of that in GloVe, we initialize their embedding weights with their corresponding $300$-dimensional pre-trained GloVe embeddings. \cite{Pennington2014GloVeGV}.
Embeddings of tokens that do not exist in GloVe are initialized with a random vector sampled from the uniform distribution $\mathcal{U}(-0.25, 0.25)$. 
We train our models on a single 32GB Nvidia V100 GPU.
Further implementation details about batch size, learning rate etc. can be found in Appendix: Table \ref{tbl:model_stats}.

\noindent \textbf{Datasets:} We evaluate the proposed framework and training pipeline on 4 publicly available benchmarks (Table: 1) from the extreme classification repository \citep{Bhatia16}.
Across the datasets, the number of labels range from 350K to 2.8 Million.

\noindent \textbf{Evaluation Metrics}
For evaluation of the methods, we use precision at $k$ (denoted $P@k$), and its propensity scored variant (denoted $PSP@k$) \cite{Jain16, Bhatia16}. 
While both aim to measure the performance amongst the top-k slots, which is a common objective for recommendation systems, the $PSP@k$ metric is more aimed at evaluating model performance on tail labels despite its shortcomings \cite{schultheis2022missing}.

\subsection{Main Results}
\textsc{SyncXML}(which is evaluated with the InceptionXML encoder, unless otherwise mentioned) finds a sweet-spot (Fig. \ref{fig:efficient}) between the two extreme ends of modern deep extreme classification pipelines - heavy transformer-based methods, and frugal architectures such as \textsc{Astec}. We show that replacing the pre-trained transformer encoder with our lightweight CNN-based encoder, combined with further improvements to the hard-negative mining pipeline leads to better prediction performance apart from faster training and the ability to scale to millions of labels. 
As shown in \autoref{tbl:results}, the \textsc{InceptionXML} encoder, outperforms the previous state-of-the-art \textsc{Astec} for all but one dataset-metric combinations.
Further, in many cases, a single \textsc{InceptionXML} model also outperforms an ensemble version \textsc{Astec-3} with non-trivial gains.

\begin{itemize}[leftmargin=*,align=left]
    \setlength\itemsep{0em}
    \item We observe at least 10\% improvement over \textsc{XML-CNN} \cite{Liu17}, which uses convolutions to capture $n$-grams for representation learning, showing the effectiveness of our approach as compared to conventional usage of convolutions on short-text queries.
    
    \item Significant gains of up to $20\%$ can be observed as compared to the transformer-based \textsc{APLC-XLNet} \cite{Ye20} and the RNN-based AttentionXML \cite{you18}. 
    
    \item \textsc{InceptionXML} also outperforms \textsc{LightXML}, which uses dynamic hard-negative mining, on all benchmarks despite having a significantly lighter architecture.
    Notably, both transformer-based approaches i.e. LightXML and APLC-XLNet don't scale to AmazonTitles-3M dataset, demonstrating the efficacy and scalability of the proposed light-weight encoder \textsc{InceptionXML} and \textsc{SyncXML} training pipeline.
    
    \item Our models also significantly outperform non-deep learning approaches which use bag-of-words representations such as the label-tree based algorithms like \textsc{Bonsai} \cite{Khandagale19} and \textsc{Parabel} \cite{Prabhu18b}, and \textsc{DiSMEC} \cite{Babbar17} which is an embarrassingly parallel implementation of LIBLINEAR \cite{fan2008liblinear}.
    \item It maybe noted that \textsc{InceptionXML} outperforms its scalable counterpart on all benchmarks, especially on PSP metrics. While \textsc{InceptionXML} receives gradient from all negative labels instead of only hard-negatives during training, it also predicts over the entire label space. On the other hand, \textsc{SyncXML} has to rely on the meta-classifier for label shortlisting. As 100\% recall cannot be ensured for label-shortlisting, some positive label-clusters are occasionally missed resulting in  slightly reduced performance. 
\end{itemize}

\subsection{Ablation Results}

\begin{figure}[!t]
    \centering
    \includegraphics[width=0.98\columnwidth]{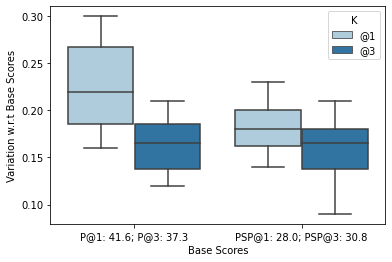}
    \caption{
    The boxplot shows a variation in performance metrics across 10 runs by randomly shuffling embedding dimensions before start of training for AmazonTitles-670K dataset. Different scores and statistics can be obtained by adding the values in the y-axis to the base scores on the x-axis.}
    \label{fig:boxplot}
\end{figure}

\noindent\textbf{Impact of Permuting Embedding Dimensions:} To show that \textsc{InceptionXML} is independent of the initial order of embedding dimensions, we randomly permute the embedding dimensions before start of the training, train with this fixed permuted order and evaluate performance. This experiment is repeated 10 times and only a \textit{slight} variation in performance metrics can be observed in Fig: \ref{fig:boxplot}. This implies that the order of embedding dimensions has \textit{little or no} impact over the results of our model.

\begin{figure*}[t]
    \centering
    \includegraphics[width=0.89\textwidth]{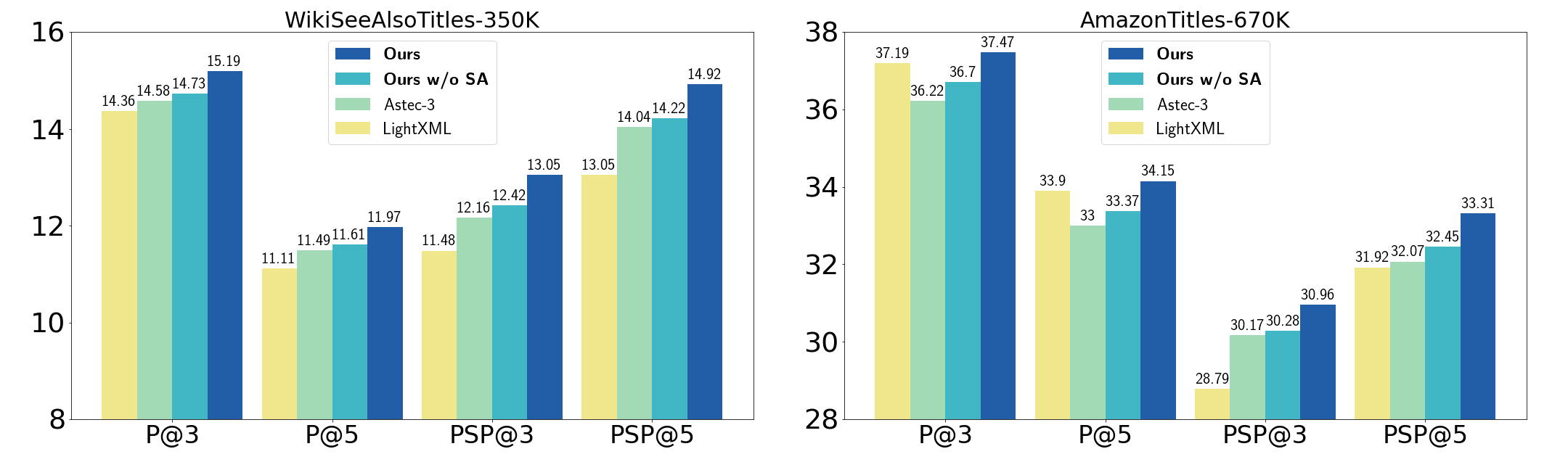}
    \caption{Performance with and w/o the self-attention layers on WikiSeeAlsoTitles-350K \& AmazonTitles-670K }
    \label{fig:SA-ablation}
\end{figure*}
\begin{table}[h]
\begin{adjustbox}{width=\columnwidth,center}
\begin{tabular}{c|cc}
\toprule
 Test data & P@3 & P@5   \\
\toprule
 Original AmazonTitles-670K  & 37.04 & 33.92   \\
 Permuted Word-order & 36.01 $\pm$ 0.05 & 32.86 $\pm$ 0.03 \\
\midrule
  Original WikiSeeAlsoTitles-350K & 14.61 & 11.44  \\
  Permuted Word-order & 14.45 $\pm$ 0.03 & 11.32 $\pm$ 0.02  \\
\bottomrule
\end{tabular}
\end{adjustbox}
\caption{Comparison of results with original test data and that obtained by permuting the word order in the test set for a single \textsc{SyncXML} model.} 
\label{tbl:word-order}
\end{table}

\noindent \textbf{Robustness to Lack of Word-order:} We train \textsc{SyncXML} on the original training data for AmazonTitles-670K, but randomly permute the words in test set, and evaluate the performance. For 10 permutations, we witness only a minor dip in performance (\autoref{tbl:word-order}), still outperforming \textsc{Astec-3}, and demonstrating the robustness of our encoder to lack of structure in short-text queries.

\begin{table}[!h]
\begin{adjustbox}{width=\columnwidth,center}
\begin{tabular}{c|c|cc|cc}
\toprule
($L', Top_K$)  & Model & P@1 & P@5 & PSP@1 & PSP5  \\
\toprule
{\multirow{3}{*}{8K, 100}}   & Ours & 40.26 & 32.75 & 26.05 & 31.21  \\
{} & Ours w/o Detaching & 40.13 & 32.68 & 25.75 & 31.07 \\
{} & in LightXML Framework & 39.40 & 32.36 & 25.14 & 30.38 \\
\midrule
{\multirow{3}{*}{16K, 200}}  & Ours & 40.67 & 33.27 & 26.34 & 31.81 \\
{} & Ours w/o Detaching & 40.51 & 32.95 & 26.03 & 31.36 \\
{} & in LightXML Framework &39.47 & 32.43 & 24.89 & 30.67 \\
\midrule
{\multirow{3}{*}{32K, 400}}  & Ours & 41.01 & 33.65 & 26.75 & 32.32 \\
{} & Ours w/o Detaching & 40.24 & 33.09 & 26.07 & 31.67 \\
{} & in LightXML Framework & 39.58 & 34.81 & 24.45 & 30.73 \\
\midrule
{\multirow{3}{*}{65K, 800}} & Ours & \textbf{41.51} & \textbf{34.39} & 27.25 & \textbf{32.89} \\
{} & Ours w/o Detaching & 40.47 & 33.23 & 26.40 & 31.97  \\
{} & in LightXML Framework & 39.27 & 32.77 & 23.54 & 30.54 \\
\midrule
\_, \_ & in DeepXML Pipeline & 38.53 & 32.21 & \textbf{27.80} & 31.62 \\
\bottomrule
\end{tabular}
\end{adjustbox}
\caption{Impact of increasing fan-out of label clusters ($L'$) on SyncXML Training Pipeline (\textit{Ours}) and LightXML Framework over AmazonTitles-670K.}
\label{tbl:ablation}
\end{table}

\noindent \textbf{SyncXML:} Table \ref{tbl:ablation} shows a comparison of the proposed \textsc{SyncXML} pipeline vis-\`a-vis \textsc{LightXML} for AmazonTitles-670K dataset. 
It is clear that the \textsc{SyncXML} training pipeline significantly improves upon the hard-negative mining technique as proposed in \textsc{LightXML}. 
Note that we keep the shortlist size consistent by doubling the number of shortlisted meta-labels as the fan-out doubles.
We also observe that, as the fan-out increases, our detached training schedule improves the results more prominently. This can be attributed to the increased ``extremeness'' of the meta-task which ensures that the 
representations learnt by the encoder for the meta-task become increasingly more relevant to the extreme-task when the gradients of the extreme classifier are re-attached during training. 

\noindent \textbf{Self-Attention Layers:} The sequentially applied self-attention layers improve \textsc{InceptionXML}'s performance by only $1\%$ at max on the performance metrics as shown in Fig.  \ref{fig:SA-ablation}. 
This further demonstrates the superior representation learning capability of our encoder for short-text queries as even without the self-attention layers, our model outperforms the ensemble model \textsc{Astec}-3 and the transformer model \textsc{LightXML}. \\

\noindent \textbf{InceptionXML in DeepXML Framework:} We integrate our encoder with the DeepXML \cite{Dahiya21} pipeline as used by \textsc{Astec} and find it inflexible to improve upon due to the requirement of fixed representations for their label shortlisting strategy. 
Moreover, when using our encoder as a drop-in replacement, we find our encoder's performance degrades in terms of precision in the DeepXML Framework as compared to the performance in the vanilla \textsc{LightXML} Framework 
(\autoref{tbl:ablation}: last row). 
This indicates the overall advantage of using dynamic hard-negative mining as compared to techniques requiring fixed representations.
\begin{table}[!h]
\begin{adjustbox}{width=\columnwidth,center}
\begin{tabular}{c|ccc|ccc}
\toprule
\textbf{Model} & \textbf{\# epochs} & \textbf{Batch Size} & \textbf{$lr_{max}$} & \textbf{$L'$} & \textbf{$Top_K$} & \textbf{ALpC}\\

\specialrule{0.70pt}{0.4ex}{0.65ex}
& \multicolumn{6}{c}{\textbf{AmazonTitles-670K}} \\
\midrule
\textsc{InceptionXML} & 42 & 128 & 0.005 & - & - & - \\
\textsc{SyncXML} & 35 & 256 & 0.008 & 65536 & 800 & 11\\
\specialrule{0.70pt}{0.4ex}{0.65ex}
& \multicolumn{6}{c}{\textbf{WikiSeeAlsoTitles-350K}} \\
\midrule
\textsc{InceptionXML} & 42 & 128 & 0.005 & - & - & - \\
\textsc{SyncXML} & 30 & 256 & 0.008 & 32768 & 800 & 10\\

\specialrule{0.70pt}{0.4ex}{0.65ex}
& \multicolumn{6}{c}{\textbf{WikiTitles-500K}} \\
\midrule 
\textsc{InceptionXML} & 33 & 128 & 0.005 & - & - & - \\
\textsc{SyncXML} & 27 & 256 & 0.008 & 32768 & 800 & 8\\

\specialrule{0.70pt}{0.4ex}{0.65ex}
& \multicolumn{6}{c}{\textbf{AmazonTitles-3M}} \\
\midrule
\textsc{InceptionXML} & - & - & - & - & - & - \\
\textsc{SyncXML} & 35 & 128 & 0.008 & 131072 & 800 & 22\\
 \bottomrule
\end{tabular}
\end{adjustbox}

\caption{Hyperparameters for proposed architectures. For \textsc{SyncXML}, $L'$ and $Top_K$ denote the number of label-clusters and the number of clusters shortlisted per dataset while ALpC denotes the average labels per cluster.}
\label{tbl:model_stats}
\end{table}

\begin{table*}
\begin{minipage}{0.25\textwidth}
    \centering
    \begin{adjustbox}{width=\textwidth}
    \begin{tabular}{c|ccc}
    \toprule
    BERT-variant & P@1 & P@3 & P@5 \\
    \midrule
        $L_2-H_{512}$ & 28.37 & 24.83 & 22.28 \\
        $L_4-H_{512}$ & 27.03 & 23.75 & 21.28 \\
        $L_2-H_{768}$ & 32.84 & 29.11 & 26.37 \\
        $L_4-H_{768}$ & 32.58 & 28.88 & 26.10 \\ 
    \midrule
    LightXML & 41.57 &  37.19 & 33.90\\
    InceptionXML & \textbf{41.51} & \textbf{37.41} & \textbf{34.39} \\
    \bottomrule
    \end{tabular}
    \end{adjustbox}
    \caption{Results using variants of BERT models with lower computational costs.} 
    \label{tbl:BERT_size}
\end{minipage}\hfill
\begin{minipage}{0.75\textwidth}
    \centering
    \begin{adjustbox}{width=0.9\textwidth}
    \begin{tabular}{c|ccc|ccc|ccc|ccc}
    \toprule
    \textbf{Encoder} & \textbf{P@1} & \textbf{P@3} & \textbf{P@5} & \textbf{PSP@1} & \textbf{PSP@3} & \textbf{PSP@5} & \textbf{P@1} & \textbf{P@3} & \textbf{P@5} & \textbf{PSP@1} & \textbf{PSP@3} & \textbf{PSP@5}\\
    \specialrule{0.70pt}{0.4ex}{0.65ex}
    & \multicolumn{6}{c|}{\textbf{AmazonTitles-670K}} & \multicolumn{6}{c}{\textbf{WikISeeAlsoTitles-350K}}\\
    \midrule
    \textsc{InceptionXML} & \textbf{41.51} & \textbf{37.41} & \textbf{34.39} & \textbf{27.25} & \textbf{30.23} & \textbf{32.89} & 20.77 & 14.61 & 11.44 & \underline{10.26} & 12.41 & 14.15\\
    \textsc{Astec} & 37.86 & 33.98 & 31.11 & 24.54 & 27.18 & 29.48 & 18.00 & 12.59 & 9.93 & 8.25 & 10.10 & 11.70 \\
    \textsc{DistilBERT} & \underline{40.87} & \underline{36.70} & \underline{33.65} & \underline{25.54} & \underline{29.28} & \underline{32.40} & \textbf{21.84} & \textbf{15.46} & \textbf{12.18} & \textbf{10.49} & \textbf{13.10} & \textbf{15.20} \\
    \textsc{DistilBERT-NoPos} & 40.22 & 36.20 & 33.22 & 25.23 & 28.92 & 32.00 & \underline{21.29} & \underline{15.06} & \underline{11.91} & 10.23 & \underline{12.75} & \underline{14.83} \\
    \specialrule{0.70pt}{0.4ex}{0.65ex}

    & \multicolumn{6}{c|}{\textbf{AmazonTitles-3M}} & \multicolumn{6}{c}{\textbf{WikiTitles-500K}} \\
    \midrule 
    \textsc{InceptionXML} & \textbf{45.64} & \textbf{44.17} & \textbf{41.35} & \textbf{14.99} & \textbf{18.26} & \textbf{19.95} & 45.24 & \underline{25.91} & \textbf{18.36} & \textbf{19.24} & \textbf{19.38} & \textbf{19.50} \\
    \textsc{Astec} & 44.56 & 41.39 & 39.89 & 13.01 & 15.92 & 17.90 & 42.73 & 23.16 & 16.35 & 16.39 & 16.08 & 16.24 \\
    \textsc{DistilBERT} & \underline{45.01} & \underline{43.86} & \underline{40.89} &  \underline{14.51} & \underline{17.87} &  \underline{19.20} & \textbf{46.93} & \textbf{25.92} & \underline{18.29} & \underline{18.32} & \underline{18.53} & \underline{18.71} \\
    \textsc{DistilBERT-NoPos} & 44.76 & 43.57 & 40.35 & 14.23 & 17.59 & 18.84 & \underline{46.08} & 25.34 & 17.88 & 17.89 & 18.08 & 18.26 \\
    \specialrule{0.70pt}{0.4ex}{0.65ex}
    \end{tabular}
    \end{adjustbox}
    \caption{Ablation results by changing the encoder in the \textsc{SyncXML} training pipeline.}
    \label{tbl:single_results}
  \end{minipage}
\end{table*}

\section{Comparison with Transformers}

\noindent In this section we discuss a vis-a-vis empirical comparison between InceptionXML and various transformers encoders in the \textsc{SyncXML} pipeline (ref: \autoref{tbl:BERT_size} and \autoref{tbl:single_results}).
\begin{itemize}
    \item Notably InceptionXML outperforms DistilBERT as an encoder on the Amazon datasets. The slightly better results of DistilBERT on the Wikipedia datasets can be attributed to the Wikipedia pre-training of BERT models. 
    \item Further, a 6 layer DistilBERT encoder in \textsc{SyncXML} pipeline outperforms the larger BERT-base in LightXML on WikiTitles-datasets while paralleling the performance on AmazonTitles datasets, thus showing the superiority of the SyncMXL training pipeline.
    \item We also investigate results on smaller BERT models formed by distillation \cite{well_read_students} to have a more comparable comparison with InceptionXML in terms of parameter count. It can be witnessed in \autoref{tbl:BERT_size} that InceptionXML significantly outperforms lightweight BERT models.
    \item To validate the lack of word-order in short-text queries, we train DistilBERT-NoPos. Since it is impractical to pre-train a BERT model from scratch without positional embeddings, we freeze and zero them out. We attain similar results as DistilBERT which validates our claim.
    \item It is worth noting that the CNN-based InceptionXML outperforms LightXML and APLC-XLNet, which use the larger BERT-base and XLNet respectively.
\end{itemize}

\subsection{Computational Cost}
\noindent \textbf{I. Training time}: As shown in Table \ref{tbl:computation_stats}, training time of \textsc{SyncXML} ranges from 4.3 hours on WikiSeeAlsoTitles-350K \& AmazonTitles-670K datasets to 27.2 hours on AmazonTitles-3M. We observe a \textasciitilde$44\%$ decrement in training time by scaling our encoder in the \textsc{SyncXML} training pipeline as compared to the unscaled \textsc{InceptionXML}. As expected, our models train much faster than transformer baselines (\textsc{LightXML}, \textsc{APLC-XLNet}) while being comparable with \textsc{Astec}.

\noindent \textbf{II. Model Size}: Notably, \textsc{InceptionXML} is extremely lightweight in terms of model size containing only 400K parameters while \textsc{SyncXML} contains only 630K parameters. This is multiple orders of magnitude lesser compared to pretrained transformer based models with \textasciitilde110 million parameters. As mentioned previously, our investigation also shows \textsc{InceptionXML} significantly outperforms lighter BERT models (\autoref{tbl:BERT_size}). 
Further, our models are approximately $8$-$10$x smaller compared to \textsc{Astec} which needs to store ANNS graphs for label centroids and training data-points for performance leading to exceptionally large model size (Table \ref{tbl:computation_stats}). 
\begin{table}[!ht]
\begin{adjustbox}{width=\columnwidth,center}
\begin{tabular}{c|cccc}
\toprule
 & \textbf{Giga} & \textbf{Training } & \textbf{Inference } & \textbf{Model } \\
 & \textbf{FLOPS} & \textbf{ Time (hr)} & \textbf{Time (msec)} & \textbf{Size (GB)} \\
\specialrule{0.70pt}{0.4ex}{0.65ex}
\textbf{Method}& \multicolumn{4}{c}{\textbf{AmazonTitles-670K}} \\
\midrule
\textsc{SyncXML} & \textbf{0.049} & 4.3 & \textbf{4.67} & 1.4\\
\textsc{InceptionXML} & 0.334 & 7.7 & 7.97 & \textbf{1.3}\\
    \textsc{Astec}  & - & \textbf{3.0} & 8.17 & 10.9 \\
\textsc{LightXML} & 2.775 & 13.5 & 13.11 & 1.6 \\
\textsc{APLC-XLNet} & 3.564 & 16.1 & 24.66 & 1.8  \\
\textsc{DistilBERT} & 0.327 & 5.7 & 6.60 & 1.4\\

\specialrule{0.70pt}{0.4ex}{0.65ex}
& \multicolumn{4}{c}{\textbf{WikiSeeAlsoTitles-350K}} \\
\midrule
\textsc{SyncXML} & \textbf{0.027} & 4.3 & \textbf{4.36} & 0.8  \\
\textsc{InceptionXML} & 0.176 & 6.3 & 5.60 & \textbf{0.7} \\
\textsc{Astec}  & - & \textbf{2.9} & 9.70 & 7.4 \\  
\textsc{LightXML} & 1.214 & 14.5 & 10.35 & 1.0 \\
\textsc{APLC-XLNet} & 2.248 & 16.0 & 18.39 & 1.5\\
\textsc{DistilBERT} & 0.137 & 5.6 & 6.27 & 1.3\\

\specialrule{0.70pt}{0.4ex}{0.65ex}
& \multicolumn{4}{c}{\textbf{WikiTitles-500K}} \\
\midrule 
\textsc{SyncXML} & \textbf{0.029} & 10.3 & \textbf{4.37} & 1.2 \\
\textsc{InceptionXML} & 0.247 & 13.5 & 6.45 & \textbf{1.1}\\
\textsc{Astec} & - & \textbf{7.3} & 9.97 & 15.1\\
\textsc{LightXML} & 1.742 & 22.4 & 11.33 & 1.5 \\
\textsc{APLC-XLNet} & 2.510 & 25.1 & 20.25 & 1.9 \\
\textsc{DistilBERT} & 0.147 & 15.3 & 6.48 & 1.5\\
 \bottomrule
\end{tabular}
\end{adjustbox}

\caption{Comparison of algorithms in terms of Giga-Flops, Training/Inference Time and Model Size. `\textbf{-}' Due to the external ANNS module used in \textsc{Astec} for label shortlisting, it is not possible to compute its flops.}
\label{tbl:computation_stats}
\end{table}

\noindent \textbf{III. Flops}: To compute flops, we use the \textsc{fvcore} library\footnote{\url{https://github.com/facebookresearch/fvcore/blob/main/docs/flop_count.md}} from facebook. Notably, \textsc{SyncXML} performs favourably with \textsc{LightXML} while requiring less than 2\% flops on average, and \textsc{InceptionXML} significantly outperforms the same with $1/8$x flops (Table: \ref{tbl:computation_stats}).

\noindent \textbf{IV. Inference Time}: Inference time has been calculated considering a batch size of 1 on a single Nvidia 32GB V100 GPU. We note that our proposed encoder not only exhibit the fastest inference times across all datasets, but also our scalable training pipeline \textsc{SyncXML} reduces the inference time to half as compared to the previous fastest \textsc{Astec}. In contrast, using transformer based approaches results in 3-5x slower inference as compared to \textsc{SyncXML}.

\noindent To summarize, our encoder improve by approximately an order of magnitude on both model size and floating point operations compared to recent baselines. 
They are also economical in terms of training time with fast inference speed, all while achieving state-of-the-art performance on all important metrics and benchmarks.

\section{Conclusion}
In this work, we develop a lightweight CNN-based encoder for the task of short-text extreme classification. 
Augmented with a self-attention based word-order agnostic module, the proposed encoder pushes the state-of-the-art across all popular benchmark datasets.
By synchronizing the training of extreme and meta-classifiers, we make improvements to the label hard-negative mining pipeline and develop a training pipeline \textsc{SyncXML} that scales our encoder to dataset with million of labels.
Importantly, these capabilities are achieved while being computationally inexpensive during training, inference, and in model size. 
As part of our future work we are looking into incorporating label text into the InceptionXML architecture, making it applicable to more problems in the domain.

\section{Acknowledgments}
We acknowledge the support from the Academy of Finland projects : 347707 and 348215, the computational resources provided by the Aalto Science-IT project, and CSC – IT Center for Science, Finland.
\newpage

\bibliographystyle{acl_natbib}
\balance
\bibliography{anthology}

\newpage
\appendix
\section{\textsc{InceptionXML-LF}}
\begin{figure*}[h!]
    \centering
    \includegraphics[width=\textwidth]{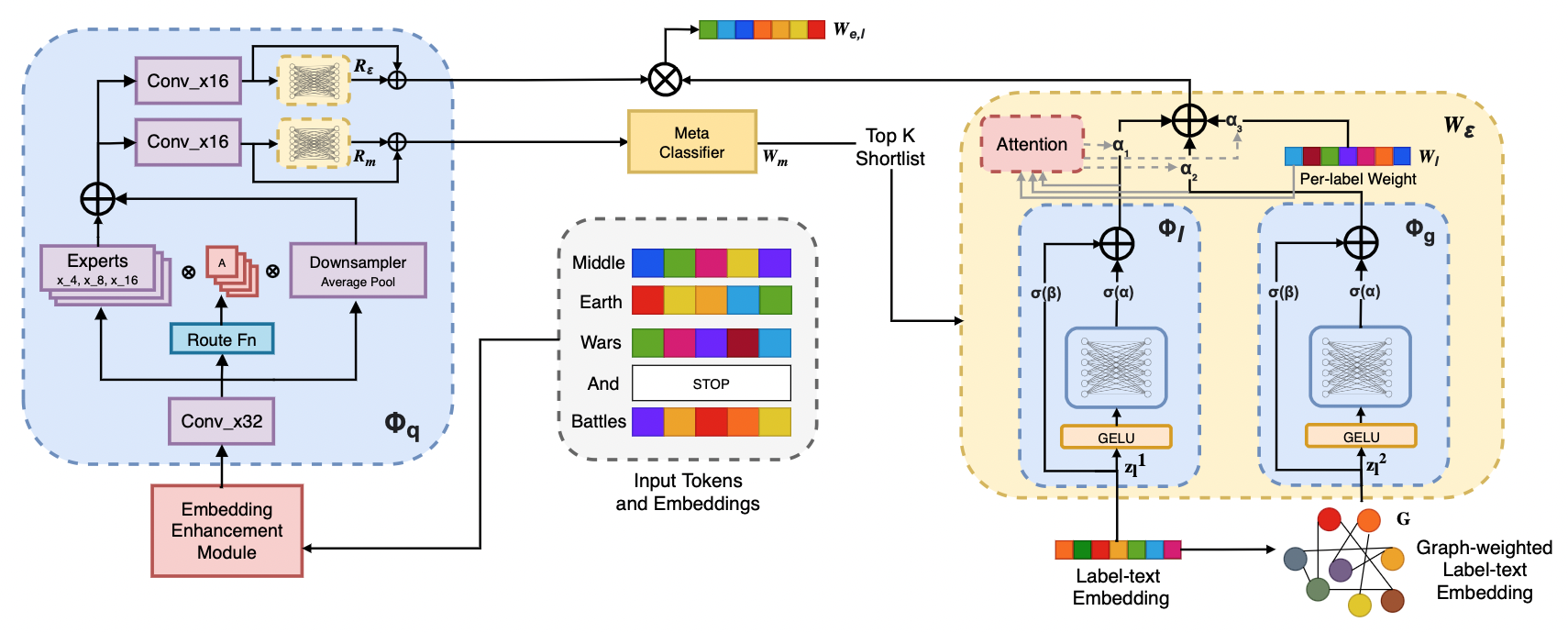}
    \caption{\textsc{InceptionXML-LF}. The improved Inception Module along with instance attention is shown in detail. Changes to the \textsc{InceptionXML} framework using the \textsc{ECLARE} classifier are also shown.}
    \label{fig:model-lf}
\end{figure*}
\textbf{Model Outlook: } Short-text queries are encoded by a modified InceptionXML encoder, which encodes an input query $x_i$ using an encoder $\Phi_q:=(E, \theta)$ parameterised by $E$ and $\theta$, where $E$ denotes a $D$-dimensional embedding layer 
and $\theta$ denotes the parameters of the embedding enhancement and the inception module respectively.
Alongside $\Phi_q$, \textsc{InceptionXML-LF} learns two frugal \textsc{Astec}-like \cite{Dahiya21} encoders, one each as a label-text encoder $\Phi_l:=\{E, \mathcal{R}\}$ and a graph augmented encoder $\Phi_g:=\{E, \mathcal{R}\}$. 
Here, $\mathcal{R}$ denotes the parameters of a fully connected layer bounded by a spectral norm and the embedding layer $E$ is shared between all $\Phi_q, \Phi_l$ and $\Phi_g$ for joint query-label word embedding learning. 
Further, an attention module $\mathcal{A}$, meta-classifier $W_m$ and an extreme classifier $W_e$ are also learnt together with the encoders. 
Next, we specify the details of all components of \textsc{InceptionXML-LF}.
\label{inceptionxml-lf}

\subsection{Instance-Attention in Query Encoder} 
We make two improvements to the inception module \textsc{InceptionXML} for better efficiency. Firstly, in the inception module, the activation maps from the first convolution layer are concatenated before passing them onto the second convolution layer. To make this more computationally efficient, we replace this ``inception-like'' setting with a ``mixture of expert'' setting \cite{CondConv}. 
Specifically, a route function is added that produces dynamic weights for each instance to perform a dynamic element-wise weighted sum of activation maps of each filter. 
Along with the three convolutional experts, we also add an average pool as a down sampling residual connection to ensure better gradient flow across the encoder. 
Second, we decouple the second convolution layer to have one each for the meta and extreme classification tasks. 

\subsection{Dynamic Hard Negative Mining}
Training one-vs-all (OvA) label classifiers becomes infeasible in the XMC setting where we have hundreds of thousands or even millions of labels.
To mitigate this problem, the final prediction or loss calculation is done on a shortlist of size $\sqrt{L}$ comprising of only hard-negatives label.
This mechanism helps reduce complexity of XMC from an intractable $O(NDL)$ to a computationally feasible $O(ND\sqrt{L})$ problem. \textsc{InceptionXML-LF} inherits the synchronized hard negative mining framework as used in the \textsc{InceptionXML}.
Specifically, the encoded meta representation is passed through the meta-classifier which predicts the top-K relevant label clusters per input query. All labels present in the top-K shortlisted label clusters then form the hard negative label shortlist for the extreme task.
This allows for progressively harder labels to get shortlisted per short-text query as the training proceeds and the encoder learns better representations.

\subsection{Label-text and LCG Augmented Classifiers}
\textsc{InceptionXML-LF}'s extreme classifier weight vectors $W_e$ comprise of 3 weights, as in \cite{Mittal21b}. Specifically, the weight vectors are a result of an attention-based sum of (i) label-text embeddings from $\Phi_l$, (ii) graph augmented label embeddings from $\Phi_g$ and, (iii) randomly initialized per-label independent weights $W_l$.

\noindent As shown in \autoref{fig:model-lf}, we first obtain label-text embeddings as $z_l^1 = E \cdot z_l^0$, where $z_l^0$ are the TF-IDF weights of label feature corresponding to label $l$.
Next, we use the label correlation graph $\mathbf{G}$ to create the graph-weighted label-text embeddings $z_l^2 = \sum_{m \in [L]} \mathbf{G}_{lm}\cdot z_l^0$ to capture higher order query-tail label correlations.
$z_l^1$ and $z_l^2$ are then passed into the frugal encoders $\Phi_l$ and $\Phi_g$ respectively. These encoders comprise only of a residual connection across a fully connected layer as $\alpha \cdot \mathcal{R} \cdot \mathcal{G}(\tilde{z}_l) + \beta \cdot \tilde{z}_l$, where $\tilde{z}_l = \{z_l^1, z_l^2\}$, $\mathcal{G}$ represents GELU activation and $\alpha$ and $\beta$ are learned weights. 
Finally, the per-label weight vectors for the extreme task are obtained as 
$$W_{e, l} = \mathcal{A}(z_l^1, z_l^2, W_l) = \alpha^1 \cdot z_l^1 + \alpha^2 \cdot z_l^2 + \alpha^3 \cdot W_l$$
where $\mathcal{A}$ is the attention block and $\alpha^{\{1, 2, 3\}}$ are the dynamic attention weights produced by the attention block. 

\subsection{Two-phased Training}

\noindent \textbf{Motivation:} We find there to be a mismatch in the training objectives in DeepXML-based approaches like \textsc{Astec}, \textsc{Decaf} and \textsc{Eclare} which first train their word embeddings on meta-labels in Phase I and then transfer these learnt embeddings for classification over extreme fine-grained labels in Phase III \cite{Dahiya21}. 
Thus, in our two-phased training for \textsc{InceptionXML-LF}, we keep our training objective same for both phases. 
Note that, in \textsc{InceptionXML-LF} the word embeddings are always learnt on labels instead of meta-labels or label clusters and we only augment our extreme classifier weight vectors $W_e$ with label-text embeddings and LCG weighted label embeddings. We keep the meta-classifier $W_m$ as a standard randomly initialized classification layer.

\noindent \textbf{Phase I:} In the first phase, we initialize the embedding layer $E$ with pre-trained GloVe embeddings \cite{Glove}, the residual layer $\mathcal{R}$ in $\Phi_l$ and $\Phi_g$ is initialized to identity and the rest of the model comprising of $\Phi_q, W_m$ and $\mathcal{A}$ is randomly initialized. The model is then trained end-to-end but without using free weight vectors $W_l$ in the extreme classifier $W_e$. This set up implies that $W_e$ only consists of weights tied to $E$ through $\Phi_l$ and $\Phi_g$ which allows for efficient joint learning of query-label word embeddings \cite{Mittal21} in the absence of free weight vectors.
Model training in this phase follows the \textsc{SyncXML} pipeline as described previously, without detaching any gradients to the extreme classifier for the first few epochs. 
In this phase, the final per-label score is given by: 
$$P_l = \mathcal{A}(\Phi_l(z_l^1),\ \Phi_g(z_l^2)) \cdot \Phi_q(x)$$

\noindent \textbf{Phase II:} In this phase, we first refine our clusters based on the jointly learnt word embeddings. Specifically, we recluster the labels using the dense $z_l^1$ representations instead of using their sparse PIFA representations \cite{Chang20} and subsequently reinitialize $W_m$.
We repeat the Phase I training, but this time the formulation of $W_e$ also includes $W_l$ which are initialised with the updated $z_l^1$ as well. 
Here, the final per-label score is given by: 
$$P_l = \mathcal{A}(\Phi_l(z_l^1),\ \Phi_g(z_l^2),\ W_l) \cdot \Phi_q(x)$$

\begin{table*}[!h]
\begin{adjustbox}{width=\textwidth,center}
\begin{tabular}{c|ccc|ccc|ccc|ccc}
\toprule
\textbf{Method}     & \textbf{P@1}   & \textbf{P@3}   & \textbf{P@5}   & \textbf{PSP@1}   & \textbf{PSP@3}   & \textbf{PSP@5} & \textbf{P@1}   & \textbf{P@3}   & \textbf{P@5}   & \textbf{PSP@1}   & \textbf{PSP@3}   & \textbf{PSP@5} \\
\specialrule{0.70pt}{0.4ex}{0.65ex}
& \multicolumn{6}{c|}{\textbf{LF-AmazonTitles-131K}} & \multicolumn{6}{c}{\textbf{LF-AmazonTitles-1.3M}} \\
\specialrule{0.70pt}{0.4ex}{0.65ex}
\textsc{Astec} & 37.12 & 25.20 & 18.24 & 29.22 & 34.64 & 39.49 & 48.82 & 42.62 & 38.44 & 21.47 & 25.41 & 27.86\\
\textsc{Decaf} & 38.40  & 25.84 & 18.65 & 30.85 & 36.44 & 41.42 & \textbf{50.67} & \textbf{44.49} & \textbf{40.35} & 22.07 & 26.54 & 29.30\\
\textsc{Eclare} & \underline{40.46} & \textbf{27.54} & \textbf{19.63} & \underline{33.18} & \textbf{39.55} & \underline{44.10} & \underline{50.14} & \underline{44.09} & \underline{40.00} & \underline{23.43} & \underline{27.90} & \underline{30.56}\\
\textsc{SyncXML}& 36.79 &  24.94 &  17.95 & 28.50 &  34.15 & 38.79 & 48.21 & 42.47 & 38.59 & 20.72 & 24.94 & 27.52\\
\textsc{InceptionXML-LF} & \textbf{40.74} & \underline{27.24} & \underline{19.57} & \textbf{34.52} & \underline{39.40} & \textbf{44.13} & 49.01 & 42.97  & 39.46 & \textbf{24.56}  & \textbf{28.37} & \textbf{31.67}\\
\specialrule{0.70pt}{0.4ex}{0.65ex}
    & \multicolumn{6}{c|}{\textbf{LF-WikiSeeAlsoTitles-320K}} & \multicolumn{6}{c}{\textbf{LF-WikiTitles-500K}} \\
\specialrule{0.70pt}{0.4ex}{0.65ex}
\textsc{Astec} & 22.72 & 15.12 & 11.43 & 13.69 & 15.81 & 17.50 & 44.40 & 24.69 & 17.49 & 18.31 & 18.25 & 18.56\\
\textsc{Decaf} & 25.14 & 16.90  & 12.86 & 16.73 & 18.99 & 21.01 & 44.21 & 24.64 & 17.36 & 19.29 & 19.82 & \underline{19.96}\\
\textsc{Eclare} & \textbf{29.35} & \textbf{19.83} & \textbf{15.05}  & \textbf{22.01} & \textbf{24.23} & \textbf{26.27} & 44.36 & 24.29 & 16.91 & \underline{21.58} & \underline{20.39} & 19.84\\
\textsc{SyncXML} & 23.10 & 15.54 & 11.52 & 14.15 & 16.71 & 17.39 & \underline{44.61} & \underline{24.79} & \textbf{19.52} & 18.65 & 18.70 & 18.94 \\
\textsc{InceptionXML-LF} & \underline{28.99} & \underline{19.53} & \underline{14.79}  & \underline{21.45} & \underline{23.65} & \underline{25.65} & \textbf{44.89} & \textbf{25.71} & \underline{18.23} & \textbf{23.88} & \textbf{22.58} & \textbf{22.50}\\
\bottomrule
\end{tabular}
\end{adjustbox}
\caption{Results comparing the performance of InceptionXML-LF with state-of-the-art XMC baselines leveraging label features. The best results are in \textbf{bold}, and the next best are \underline{underlined}.}
\label{tbl:InceptionXML-LF}
\end{table*}

\subsection{Empirical Performance}
\noindent \textbf{Datasets and Baselines: }We compare the performance of InceptionXML-LF against SOTA extreme classifiers Astec, DECAF, ECLARE and SyncXML. We perform all experiments on four standard short-text XMC datasets with label-features, whose details are in \autoref{tbl:lf-datasets}.

\noindent \textbf{Observations: }Notably, InceptionXML-LF, with half the number of parameters in each encoder, performs at par with ECLARE, and often outperforms it across various datasets and metrics. This further validates the efficacy of the base InceptionXML encoder. InceptionXML-LF also consistently outperforms SyncXML, demonstrating the effectiveness of the additional encoders and training strategy to leverage label features.

\begin{table}
    \centering
    \begin{adjustbox}{width = \columnwidth, center}
    \begin{tabular}{cccccc}
        \toprule
        \textbf{Datasets} & \textbf{N} & \textbf{L} & \textbf{APpL} & \textbf{ALpP} & \textbf{AWpP} \\
        \midrule
        LF-AmazonTitles-131K & 294,805 & 131,073 & 5.15 & 2.29 & 6.92\\
        LF-WikiSeeAlsoTitles-320K & 693,082 & 312,330 & 4.67 & 2.11 & 3.01\\
        LF-WikiTitles-500K  & 1,813,391 & 501,070 & 17.15 & 4.74 & 3.10\\
        LF-AmazonTitles-1.3M & 2,248,619 & 1,305,265 & 38.24 & 22.20 & 8.74\\
        \bottomrule
    \end{tabular}
    \end{adjustbox}
    \caption{Details of short-text benchmarks with label features. APpL is the avg. points per label, ALpP being avg. labels per point and AWpP is the length i.e. avg. words per point.}
    \label{tbl:lf-datasets}
\end{table}

\end{document}